# Opinion Mining for Relating Subjective Expressions and Annual Earnings in US Financial Statements

Chien-Liang Chen[†], Chao-Lin Liu[‡], Yuan-Chen Chang[§] AND Hsiang-Ping Tsai[↑]

[†, ‡]Department of Computer Science; [§]Department of Finance, [↑]College of Management
[†, ‡, §]National Chengchi University, Taiwan; [↑]Yuan-Ze University, Taiwan
{[†]98753013, [‡]chaolin, [§]yccchang}@nccu.edu.tw, [↑]hptsai@saturn.yzu.edu.tw

Financial statements contain quantitative information and manager's subjective evaluation of firm's financial status. Using information released in U.S. 10-K filings. Both qualitative and quantitative appraisals are crucial for quality financial decisions. To extract such opinioned statements from the reports, we built tagging models based on the conditional random field (CRF) techniques, considering a variety of combinations of linguistic factors including morphology, orthography, predicate-argument structure, syntax, and simple semantics. Our results show that the CRF models are reasonably effective to find opinion holders in experiments when we adopted the popular MPQA corpus for training and testing. The contribution of our paper is to identify opinion patterns in multiword expressions (MWEs) forms rather than in single word forms.

We find that the managers of corporations attempt to use more optimistic words to obfuscate negative financial performance and to accentuate the positive financial performance. Our results also show that decreasing earnings were often accompanied by ambiguous and mild statements in the reporting year and that increasing earnings were stated in assertive and positive way.

## 1. INTRODUCTION

Opinion mining and sentiment analysis have been applied to many topics in computer science and finance literature. Loughran and McDonald [20], examining textual statements and financial figures, developed positive and negative unigram lists that better reflect the tones of the U.S. financial statements in 10-K filings than words in traditional psychology dictionaries, such as Harvard-IV-4 Psychological Dictionary. Engelberg studied the mechanism behind the under-reaction of asset prices to qualitative information, e.g., dividend initiation, in relevant media with the help of typed dependency parsers [9].

Most stock investors do not have the opportunity to directly observe a firm's daily operating activities and thus rely on the financial statements to obtain information about a firm's profit-generating ability. With the information, investors can evaluate a firm's value and decide whether to invest in the stock or not. Financial statements contain two types of information: quantitative information (i.e., reported earnings) and qualitative information (i.e., the description that reveals managers' subjective evaluation of a firm's financial status) [17]. Earlier studies have documented that the reported earnings help predict a firm's future earnings and cash flows. Recently, some researchers turn their attention to whether the qualitative information (i.e., description) also helps predict a firm's future performance. These studies have reported that the language used in Internet stock message board [1], news stories [35], financial statements [16], earnings press releases, and the speech of corporate managers during conference calls may help predict a firm's future performance. Unlike previous studies that examine only individual words, our work aims to study the relationship between MWEs from financial reports and a firm's performance.

We propose a computational procedure based on conditional random field (CRF) [15] models to identify opinion holders and to extract subjective opinion patterns in financial statements. Naturally, the opinion holders and the opinion patterns are multiword expressions (MWEs). We considered an array of linguistic features including morphology, orthography, predicate-argument structure, syntax, and simple semantics. To evaluate the CRF models effectively, we trained and tested the models with annotated MPQA corpus [23].

A major contribution of our work is to automatically extract lists of subjective words that are linked to positive and negative financial statuses with subjective MWEs. Opinion patterns in financial statements include opinion holders and subjective MWEs. For instance, the opinion patterns in the sentence "The Company believes the profits could be adversely affected" include opinion holder "The Company" and two subjective expressions: "believe" and "could be adversely affected". Unlike traditional models that considered individual words (unigrams) and "bag of words" [19], we attempt to automatically extract MWEs from textual statements. MWEs capture manager's subjective evaluations of the financial status of the reporting firms more precisely, and are more informative than individual words for investors.

Previous researches investigated the relationship between financial performance and the textual content of financial press reports. Antweiler and Frank studied the influence of Internet stock message board on the stock markets using 1.5 millions messages posted on Yahoo! Finance and Raging Bull on 45 companies. A naïve-Bayes algorithm was applied to measure bullishness sentiment in these messages. Their results showed that the stock messages help predict market volatility but not stock returns [1]. Li examined the link between risk sentiment of corporations' financial statements with stock returns and future earnings. Risk sentiment of annual reports is measured by the frequencies of the individual words that are related to the risk or uncertainty in 10-K filings. Li found that the risk sentiment is negatively correlated with future earnings and future stock returns [16]. Tetlock et al. found that negative words reported in financial news regarding S&P 500 firms capture some hard-to-be-quantified fundamentals. Although the markets under-reacted to negative words for firm-specific news, the results showed that the negative words, especially those related to the firm's fundamentals, are useful predicators for low earnings and returns [35].

We applied multinomial logistic regression models to explore the link between MWEs that we extracted from the financial statements and future earnings of the reporting companies. We find that companies use more affirmative and positive terms with increasing earnings. In contrast, they use relatively mild and ambiguous wordings when earnings decline.

Although we used state-of-art techniques to extract MWEs from the financial statements, the effects of extraction are subject to limitation of previous research, e.g., [6]. Hence, we employed domain knowledge to manually select useful MWEs from the MWEs that were identified by our computer model, and examined the statistical relations between the selected MWEs and the annual earnings of the reporting companies.

Another contribution of our work is thus to reveal interesting relations between the MWEs and the earnings. For instance, when the companies face declining earnings, managers tend to include optimistic or mild phrases such as "will be successful" and "could be adversely" (rather than "will be adversely") in financial statements. When the company's earnings increase, the financial statements are more likely to contain positive and less ambiguous wordings such as "can successfully" and "reasonably assured".

We describe the procedures for preparing the financial data about U.S. companies in Section 2, discuss our methods for identifying opinion patterns with CRF models in Section 3, and present the application of multinomial logistic regression to relate the subjective MWEs with annual earnings in Section 4. Section 5 evaluates our CRF models with the MPQA corpus. Section 6 describes the opinion patterns and their combinations that were extracted from the financial statements. Section 7 investigates the relationships between financial MWEs with the annual earnings of the reporting companies. A summary of our work and contributions is provided along with discussions about some challenging issues in current work in Section 8.

## 2. PROBLEM DEFINITION AND DATA SOURCES

Our research goal is to examine relationships between the manager's subjective expressions with the annual earnings in the financial statements of U.S. listed companies. Fig. 1 presents a block diagram for the work flow of our work; the meanings of individual steps will become clear in later discussion.

We downloaded the 10-K filings (i.e., financial statements) of public companies from the EDGAR database [8] that is governed by the U.S. Securities and Exchange Commission. Opinion patterns were extracted from 2102 annual statements of 324 U.S. companies between 1996 and 2007. The original data in EDGAR were stored with the HTML format. Like typical web pages, the HTML documents available in EDGAR contain tags and some irrelevant information, so we extracted the English sentences from the footnote sections in individual filing with the procedure shown in Fig. 2, where data are shown in rectangles and processing steps are shown in rectangles with rounded corners.

We removed the HTML tags, pictures, tables, front and ending matters, and exhibitions to retrieve the useful item sections from financial statements. We then eliminated the lines that contain too many white spaces, symbols, numbers or meaningless strings by regular expressions. Finally, we adopted a LingPipe [18] sentence model to segment the filings into 1.3 million sentences. We selected 85394 sentences contained "subjective words". We used two sources of "subjective words". MPQA provides two lists of words related to "objective speech event" and "direct-subjective" (e.g., "believe"). Loughran and McDonald [19] compiled lists of financial words with positive and negative unigrams (e.g., "failure"). After this filtering, we chose 30381 sentences that contained between 8 and 100 words that could be handled by our parsers.

We obtained information about annual earnings of public companies from the Compustat database [30]. Using links provided by Sufi [32], we merged the EDGAR and Compustat data about individual companies by matching names and dates. Fig. 3 shows the matching process. After eliminating the data with missing values, the number of data instances reduced from 2102 to 1421, and we refer to this set of data as the "small dataset". To obtain data in an experiment of larger scale, we considered the financial statements of 6534 U.S. companies, that were issued between 1996 and 2007. We expanded the sample size to 22780 instances, again after dropping instances with missing values.

## 3. IDENTIFYING OPINION PATTERNS

Liu applied opinion mining and sentiment analysis techniques to identify the text or speech that carry opinions, sentiment, or emotion [19]. Text in financial statements can be classified into two types: fact discourse or opinion discourse. The former provides objective information about specific objects (e.g., entities, events, or topics). In contrast, the latter contains subjective evaluation, belief, judgment, or comments about the objects. The main difference is whether or not subjective expressions are included in the discourses. Assuming that human's subjective feelings about objects have only two extremes, i.e., pleasantness and unpleasantness, we categorize opinion expressions into three kinds of polarities: positive, negative, and neutral. For a given polarity, human beings may have different degrees of feelings about different things. For example, "wrong" and "might be inaccurate" are negative words expressing disapproval of someone or something, but the word "wrong" conveys stronger disapproval than "might be inaccurate" pragmatically.

Previous research employed different machine learning techniques to determine the sentiments in texts of different granularities. Some worked for document-level; others worked for paragraph-level, sentence-level, phrase-level, or word-level. Pang et al. utilized the concept of naïve Bayes, maximum entropy classification, and support vector machines to classify the sentiment polarity of movie reviews at the document-level [24]. Wiebe et al. classified subjective sentences based on syntactic features such as syntactic categories of the constituents [36]. Riloff and Wiebe used a bootstrapping machine learning process to extract the subjective expressions from sentences, manually annotated both the strengths and polarities of the subjective expressions, and collected them into a subjective word list [28]. Kim and Hovy used syntactic features to indentify the opinion holders in the MPQA corpus by a ranking algorithm that considered maximum entropy [14]. Choi et al. adopted a hybrid approach that combined conditional random field and the AutoSlog information extraction learning algorithm to identify sources of opinions in the MPQA corpus [6].

Due to the availability of the MPQA corpus, we employed the corpus [23] to train and evaluate our CRF models for opinion mining, and then utilized the CRF models to identify opinion holders and opinion patterns in the financial statements. We did not use the financial statements to train the CRF because training the CRF models is supervised learning, which required annotated data. However, there is no known source of financial statements with appropriate annotations yet. MPQA is a corpus with the annotation that we needed, so was chosen for our work.

A number of previous researches on opinion mining used the MPQA corpus for semantic labels. The MPQA corpus contains different topics of news that were collected from different sources. Since the annotation unit of MPQA is at the sentence level, we also focus on the opinion mining at the same level.

We used the MPQA corpus to train CRF models for identifying opinion patterns, and considered class labels for tagging five different aspects of opinions: "agent", "expressive-subjectivity", "objective speech event", "direct-subjective", and "target" (cf. [23] for detail definitions).

Proponents of the linking theory argue that prediction of semantic role labeling by syntactic information and predicate-argument structure is feasible [11]. Because the identification of opinion holders is an instance of semantic role labeling, we believe that it is reasonable to use general linguistic features for opinion holder identification.

To better identify opinion holders, we annotated our data with the IOB format [27], which was widely employed in NP chunking, word segmentation, and named entity recognition research. Table 1 provides a sample sentence and its annotations in the MPQA corpus. It is seen that, "according to" is tagged as "B-objective speech event" and "I-objective speech event" in sequence, where "B" stands for the first word of a phrase and "I" stands for an internal word of a phrase. The word "believe" is both the first and the internal word of a segment, and is tagged as "B-direct-subjective". The labels are not overlapped and mutually embedded. We defined the opinion holders as a phrase with label "agent" in the corpus. Opinion holders do not have to be persons; they can be any entities that express opinions, beliefs, and speculations.

Inheriting from the choices in MPQA, we treated an individual sentence as a training datum, and annotated the sentences with token-level, phrase-level, and sentence-level features. No contextual information that considers two or more sentences was considered. The features include morphological, orthographical, predicate-argument structure, syntactic, and simple semantic information. Table 2 provides an illustration of the features for the sentence "We decided to make some bold decisions" and shows how we created features for a given sentence. The token-level feature value of each token in a sentence is extracted and recorded sequentially, but sentence-level feature value of a sentence is processed only once.

We extracted the linguistic information with existing software available on the Internet. They include the Stanford NLP tools [31], ASSERT semantic role labeler [3, 26], and CGI Shallow parser [13] for linguistic features. We employed regular expressions to judge and determine some simple features.

In the following subsections, we list and explain the linguistic information that we obtained from the training data to build linear-chain CRF models. We will report (1) how the models were evaluated with the MPQA corpus in Section 5 and (2) how the best CRF models were applied to extract the subjective expressions in the financial statement in Section 6.

### 3.1 Morphological and Orthographical Features

*Original tokens* (**f1**): we tokenized the words in a sentence by white spaces and punctuations, and used them as a feature.

*Lemmatize tokens* (**f2**): the original token was lemmatized by the Stanford parser.

We employed regular expressions to judge the true values for the third, fourth, and fifth features. We call these three features as *orthographical* features in the experiments.

*Initial words, words in capital letters or with first character capitalized* (**f3**): in English, abbreviations of words or words all in capital letters are probably all or part of the name of specific entities.

*Words with alphabetic letters and mixing numbers* (**f4**): it is observed that some organizations tend to have a name with alphabetic letters and mixing numbers for being ease to memorize (e.g., the U.S. company "3M").

*Punctuations* (**f5**): punctuations function well in marking boundaries of semantic units and separate different phrases or clauses. Quotation marks often indicate the sentences of speech or subjective expressions.

### 3.2 Predicate-Argument Structure Features

The predicate-argument structure (PAS) has been implemented in labeling semantic roles [34]. PASes are structures that can be used to capture the events of interest and the participating entities involved in events that correspond to predicates and arguments, respectively. Generally speaking, a predicate is about a verb that conveys the type of events. The types and numbers of arguments vary when the predicates differ in syntactical or pragmatic views (e.g., transitive vs. intransitive). We relied on ASSERT to extract the following information [3].

*Position of predicate* (**f6**): arguments are usually close to the verbs, especially agents and patients (subjects and objects of verbs).

*Before or after predicate* (**f7**): the arguments before or after the predicate perform dif-ferent types of semantic roles. For example, in the sentence "Peter chases John", since the verb is "chases" and two arguments, "Peter" and "John', correspond to agent (subject) and patient (object), respectively. The relative positions of the arguments from the verb have influences on the semantic roles.

*Voice of the main verb* (**f8**): whether the predicate is an active or a passive voice can affect the type of arguments. In the sentence "John is chased by Peter", the verb is in its passive voice, and positions of the arguments are changed. Considering relative positions around the verb (i.e., f7) and the voice of the verb (i.e., f8) makes the resolution of the opinion holders more feasible.

### 3.3 Syntactic Features

*Sub-categorization of main verbs* (**f9**): this feature is related to the sub-structures of the main verb phrases (VPs). The dashed box in Fig. 4 indicates that the sub-categorization of "decided" is "VP➔VBD-S". This feature is useful for the analysis of the phrases or the clauses that follow the predicate, and it increases the ability in discriminating between arguments.

*Head words and their POS tags* (**f10**): the features are the head word of the phrase and the syntactic category of the head word. The number 2 dashed link in Fig. 4 shows that "decision" is the head word for the NP. Different head words in noun phrases can be used to express different semantic roles. A head word like "he" or "Bill" rather than "computer", is more likely for a noun phrase to serve as an opinion holder. Collins' head word algorithm is adopted to recognize head words [7]. Since the head of prepositional phrase (PP) is preposition and the significant semantic meaning in PP might be the noun phrase (NP), we also added the head of NP in PP as a feature which is named as content word of PP.

*Syntactic category of phrase* (phrase type, **f11**): different semantic roles tend to be realized by different syntactic categories. The opinion holders are usually noun phrases, and subjective expressions ("objective speech event" and "direct-subjective") tend to be verb phrases. In addition to using the head words and the POS tags of phrases as features, we tried to track the path from the head words to the phrases, and collected syntactic information on the way accordingly. Consider "make" in Fig. 4 as an example. It is the head word for some VPs. Tracking three levels upwards from "make", we would see VB, VP, and VP. The word "make" is the head words in all of these structures, so we would consider VB, VP, and VP features of "make". Take "decisions" as another example (cf. number 2 dashed link in Fig. 4). Again, tracking upwards from "decision" three levels, NNS, NP, and VP, we treat "decisions" as the head word for NNS and NP but not for VP. Hence, only NNS and NP became features for "decisions." We limited the levels to trace upwards to three to confine the resulting space complexity. We hope that providing contextual information will help our classification models to make better decisions.

*Syntactic path and partial path* (**f12**): according to Gidea's statistical results [11], the path VB↑VP↓PP has 14.2% chance to be PP argument or adjunct; path VB↑VP↑S↓NP↓ is 11.8% to be subject; path VBD↑VP↑NP↓ has 10.1% chance to be object of a certain sentence; and the VB↑VP↓ADVP is adverbial adjunct with 4.1%. Gidea's findings indicate that path information helps predict semantic labels. The syntactic path feature describes the syntactic relation from individual constituents to the predicate in the sentence with the syntactic categories of the traversed nodes. In Fig. 4, the number 1 dashed link from "We" to "decides" can be represented as either "PRP↑NP↑S↓VP↓VBD" or "NP↑S↓VP↓VBD" depending on whether the constituent is PRP or NP of word "We". A deep parse tree can make the complete string of path too long, which results in data sparseness problem. The partial path is part of syntactic path which contains the lowest common ancestor of constituent and predicate. (e.g., the lowest common ancestor is S in the sentence, so the partial path is reduced to "PRP↑NP↑S") Using the partial path feature can alleviate the sparseness problem.

*Base chunk* (**f13**): the base chunk feature is similar to the phrase type feature but without the phrase type overlapped. The sentence S consists of NP (We) and VP (decided to…), but this VP can be divided into non-overlapping sub-phrases which are combined by VP (decided to make) and NP (some bold decisions). We represent the base chunk in IOB format which makes the segmentation of phrase boundary more precise.

*Subordinate noun clause after verb and noun phrase before verb phrase* (**f14**): since our phrase type feature consist of only three levels of syntactic category from the parents of parse tree leaf nodes, the macro syntactic structure information can be omitted if the parse tree is constructed deeply. In sentence "The management believed that …," the subordinate noun clause following the verb "believed" may be a subjective expression. We used the Stanford Tregex to extract such patterns from the parse trees [31].

*Syntactic dependency* (**f15**): the feature is to capture the grammatical relation that includes three types of grammar dependencies: "subject relationship" (`nsubj` in Stanford parsers), "modifying relationship" (`amod` and `advmod` in Stanford parsers) and "direct-object relationship" (`dobj` in Stanford parsers). The subject relationship includes the "nominal subject" and the "passive nominal subject", which corresponds to the noun that is the syntactic subject of active and passive clauses. The modifying relationship consists of adjectival modifier or adverbial modifier, which can be any adjectival (adverb) word that modifies a noun (verb or adjective). The number 3 dashed link in Fig. 4 provides an example of the `amod` relationship, and the number 4 dashed link shows an example of the `nsubj` relationship. The direct-object relationship indicates the noun that is the direct object of the verb. We utilized the Stanford dependency parser to obtain the dependency features [31]. The opinion holders, opinion words in subjective expressions, and opinion targets are correlated with the subject, modifying, and direct-object relationship, respectively. In Fig. 4, the label of the phrase "to make some bold decisions" is "expressive-subjectivity", and we can observe that the opinion word in the phrase is "bold" with an adjective POS that modifies the noun "decisions". Since the word "we" is the subject of verb "decides", the identification of the relationship of the subject with the verb can be used to predict the opinion holders.

### 3.4 Simple Semantic Features

*Named entities* (NER, **f16**): when utilizing the syntactic features, it is hard to distinguish the entity names from the other noun phrases. The named entities can better identify the name of a person, who may be the opinion holder or the opinion target. Stanford Named Entities Recognition (NER) tool [31] was employed to label the name of persons, organizations, and locations.

*Subjective word and its polarity* (**f17**): the subjective words appear in sentences can serve multiple functions. Subjective words are used in opinioned sentences, so they are useful for detecting the opinion words for labeling "*expressive-subjectivity*" and "*direct-subjective*".

The subjective words can be classified based on two dimensions: subjectivity and polarity. We employed the classification and word lists collected by Wilson et al. [37]. On the dimension of subjectivity, words can belong to *objective*, *weak subjective*, and *strong subjective*. If a word is subjective, either weak or strong, it can carry *negative*, *neutral*, or *positive* message. Hence, there are seven categories of words in the lists.

*Verb cluster* (**f18**): verbs with similar semantic meanings might appear together in the same document. In order to arrange the semantically-related verbs into one group, we use verb clusters to avoid the occurrence of the presence of infrequent verbs which have a negative impact on the model performance. The ASSERT toolkit adopts a probabilistic co-occurrence model to cluster the co-occurrence of verbs into 64 clusters.

*The frame of the predicate in FrameNet* (**f19**): the FrameNet [10] is a corpus of sentences that has been hand-annotated for predicates and their arguments. Predicates are grouped into semantic frames around target verbs serving as semantic roles. Since an individual verb indicates specific event of interest, the predicate-argument structure would be totally different. We used the FrameNet for querying of frame names of verbs.

## 4. RELATING SUBJECTIVE EXPRESSIONS AND ANNUAL EARNINGS

After identifying the subjective MWEs in the financial statements, we investigate whether the subjective MWEs reflect the trends of firms' earnings. We used multinomial logistic regression (*Stata* [33]) to model the relationships between annual earnings and subjective MWEs, and to infer their economic implications.

### 4.1 Dependent Variable: Standardized Unexpected Earnings

We focus on the standardized unexpected earnings (*SUE*) as a proxy of the financial status of the company (Li [16] and Tetlock et al. [35]). The *SUE* for each firm in year *t* is defined by Equations (1) and (2).

$$UE_t = E_t - E_{t-1}, t = 1, 2, 3, \ldots, 12 \qquad (1)$$

$$SUE_t = (UE_t - \mu_{UE})/\sigma_{UE} \qquad (2)$$

$E_t$ denotes the earnings of a given firm in a year *t*. $UE_t$, the unexpected earning in year *t*, is the difference between the earnings of consecutive years.      is the mean of the unexpected earnings of a firm within 12 years (1996-2007, cf. Section 2), and it represents the average variation of the firm's performance.      is the standard deviation of the unexpected earnings of a firm within 12 years, and it can be used as an indicator of the long term volatility of the firm's performance. Hence, we use $SUE_t$ as the indicator of the "surprising" unexpected performance of a firm in year *t*.

We transformed *SUE* into a categorical variable (*Y*), that has three possible values: positive (1), not changed (0), and negative (-1); and the transformation rule is described in Equation (3). We set the constant    to 0.5 or 1.0 in Section 7. We categorized $SUE_t$ to avoid our judgments affected by minor changes in $SUE_t$.

$$ \qquad (3)$$

### 4.2 Multinomial Logistic Regression

Multinomial logistic regression models allow us to predict the probability of an event. We input the training data to a logistic function to capture the conditional probability distribution         , where *Y* is the $Y_t$ in Equation (3) for the performance of a company in a particular year and              is the vector of explanatory variables (cf. Section 4.3). The function form is listed as follows [12]. If        in (4), the numerator becomes 1 on the right hand side.

$$ \qquad (4)$$

$$\text{(5)}$$

is the log odd-ratio, which assigns a value of 0 as the base category for comparison, and *c* could be either category 1 or category -1. The explanatory variable is a vector that consists of the control variables and MWEf-idf variables, while is the intercept and is the vector of the regression coefficients. A positive coefficient of the regression model indicates that the explanatory variable increases the probability of the category *c*, while a negative coefficient indicates that the explanatory variable decreases the probability of the category *c*.

### 4.3 Explanatory Variables: MWEf-idf and Control Variables

To investigate the relationships between the MWEs in financial statements and the changes in annual earnings, we quantify individual MWEs, which were used as explanatory variable *X* in Equation (4). The coefficient of an MWE thus indicates its influence on the annual earnings.

In addition to the MWEs, the *X* vector also includes other control variables that are related with the *SUEs*. These control variables are quantitative financial factors such as lag *SUE* (*SUE* of the previous year), BM ratio (natural log of dividing book value by market value), ROE (return on equity), accruals (earnings minus operating cash flow), size (natural log of market value), Dividend (cash dividend divided by book value), and Bankruptcy Score (Z-score) [30].

We adopt the basic concept of TFIDF (term frequency and inversed document frequency) from natural language processing and information retrieval [22] to quantify the contribution of individual MWEs. Rather than using the original TFIDF formula, we used a slightly revised formula for the TF and IDF components of MWEs, and referred to the results as *MWE frequency and inversed document frequency* (MWEf-idf) weighting. Equation (6) shows our formula.

$$\text{(6)}$$

is the weight of the $i^{th}$ MWE in the $j^{th}$ document. Similar to the traditional formula for TFIDF, is the frequency of the $i^{th}$ MWE that appears in the $j^{th}$ document, is the number of documents that contain the $i^{th}$ MWE, and is the number of the documents in the corpus. We substituted the total number of tokens in the traditional formula for TF with the constant , and changed the formula for IDF by multiplying by the constant . Using for document lengths helped us avoid dropping important but infrequent MWEs, and multiplying with helped us keep the very frequent MWEs. Using the tradition IDF will lead us to ignore phrases that appear in every document. For the current work, we would prefer to keep good MWEs as candidates for later inspection and selection. We set *q* to 20 and *l* to 40, based on observations in a small empirical trial.

We extracted subjective MWEs using the CRF model discussed in Section 3, and employed the software *Lucene* [2] to estimate the MWEf-idf values efficiently.

### 4.4 Strategy to Identify Discriminative MWEs

By fitting the logistic regression with the extracted MWEs and *SUEs* (Section 4.2), we examined whether MWEs were correlated with annual earnings. The process was conducted with a standard statistical procedure in Stata [33].

Previous work in sentiment analysis typically consider two polarities, i.e., positive and negative, e.g., [25]. Recently, researchers start to consider more categories of sentiments, e.g., [24]. We follow this trend and attempt to consider five levels of sentiments in our work. The finer classification is not perfectly scientific but qualitatively intuitive.

In order to identify the discriminative MWEs that are correlated with the *SUEs* in different strengths, we propose a strategy to rank the contribution of each MWE, *w*, to the conditional probability          in Table 3. The types of significance of *w*'s coefficients are represented as positively significant (+), negatively significant (-), not statistically significant (NSS). Based on the significance of the coefficients and the different signs of the coefficients for two opposite earning trends, we defined five categories of the contribution of an MWE by *R*(*w*): from 1$^{st}$ to 5$^{th}$, where the 1$^{st}$ is the most likely discriminative indicator of the *SUE* and the 5$^{th}$ the least likely. To simplify our notation, we use *w* to denote an individual MWE in *R*(*w*) in the table. The different values of *R*(*w*) represent qualitative differences in the explanatory power of the MWE about *SUEs*. We explain the contents of Table 3 next.

Goal: Determine *R*(*w*) of an MWE *w*, given the signs of the coefficients of quantified *w* in Equation (5) (        and         ) and the corresponding *p*-values which were estimated by Stata.

Step 1: If *w* does not significantly influence *Y* when         and when        , *R*(*w*) is 5$^{th}$. If *w* significantly influences *Y* when         and when         and if the coefficient of *w* has the same sign when         and when        , then *R*(*w*) is 4$^{th}$. Otherwise, go to Step 2.

Step 2: If the coefficients for         and         are significant and have different signs, *R*(*w*) is 1$^{st}$. If the influence of *w* on *Y* is significantly positive only when         but not when        , then *R*(*w*) is 2$^{nd}$. Similarly, if the influence of *w* on *Y* is significantly positive only when         but not when        , then *R*(*w*) is 2$^{nd}$. Analogously, if the influence of *w* on *Y* is significantly negative only when         but not         or only when         but not when        , then *R*(*w*) is 3$^{rd}$.

*R*(*w*) of *w* is an ordinal value that we used to rank the strength of influence of *w* on           . We differentiated the second and the third ranks based on our belief that positive relationships are more important than negative relationships; that is to say, the positive relationship suggests that an MWE *w* is a direct indicator of the *SUE*, and negative relationship implies an MWE *w* as an indirect indicator.

In Section 6, we use *R*(*w*) to measure the relationships of MWEs and *SUEs*. We provide evidences to show that values of *R*(*w*) of different MWEs indicate that manager's of the companies tend to use different strengths of positive and negative statements when reporting different annual earnings.

## 5. EVALUATNG CRF MODELS WITH MPQA

We report the results and procedures for evaluating the CRF models in this section.

### 5.1 Data Preparation and Evaluation

We prepared the training and test data from MPQA with the following steps. Firstly, we preprocessed the MPQA corpus to make sure that they were acceptable for the tools that we used to extract linguistic features. Secondly, we extracted the linguistic features that we discussed in Section 3. Thirdly, we chose some combinations of features and created data for training and testing the CRF models. Finally, we trained and evaluated the resulting CRF models.

The CRF models were implemented with the *MALLET toolkit* [21]. We trained and tested the CRF models with 10325 sentences from the MPQA corpus with a heldout test procedure. Training and testing were performed on 70% and 30% of the 10325 sentences, respectively. The training iteration was 500 and the Gaussian variance was 10 with first-order CRF models. The models were evaluated by precision (*p*), recall (*r*) and measures. The correct prediction of opinion labels was defined as an exact match between the labels predicted by our CRF models and the labels annotated in the MPQA corpus. *Precision* is the proportion of opinion labels predicted correctly by the model, and *recall* is the proportion of correct opinion labels predicted by the model. The definition measure is in Equation (7). We set to 1 to allow equal contributions of precision and recall to F, the resulting F is .

$$ \tag{7} $$

We calculated precision and recall ratios using individual phrase as the basis. More specifically, a predicted label sequence was considered as correct only if all of the tags for the tokens in a phrase were correct.

It is possible to consider the prediction accuracy (*a*) at the token level. For instance, if we predicted correctly the tags of four tokens in a phrase that contains five tokens, the accuracy (at the token level) will be 0.8 (=4/5), but the precision will be 0. Our system made five predictions at the token level and one set of decisions at the phrase level.

**5.2 Experimental Results for Different Combinations of Features**

Table 4 shows the results of experiments that we conducted for some combinations of features that we discussed in Section 3. The table is divided into two panels, based on different ways we treated the sentences in the experiments. There are long sentences that contain a main clause and a subordinate clause. In experiments reported in the upper panel, we treated such a long sentence as a sentence, but in experiments reported in the lower panel, we treated the clauses as separate sentences. Namely, a long sentence would be separated into multiple shorter sentences.

We did not find orthographical features as useful as we expected. In Table 4, the feature set B (only orthographical feature) has the lowest token accuracy, and the feature set H (orthographical feature added) has significantly decreased the recall by about 6% when comparing with feature set G. On the other hand, the orthographical feature captures those that are served by NER features and POS of token. By comparing the performances among feature sets G, H and I, we found that the inclusion of f14 (set I) achieved better performance than the inclusion of orthographical features (f3~f5, set H). Hence, we did not use the orthographical features in experiments other than B and H.

The difference between feature set C (POS) and J (phrase type) is in granularity. POS tags are syntactic features for tokens, and the phrase type tags are for larger constituents. Since the linguistic characteristics of both are similar, the slight difference between their measures is not surprising. Although their recalls are unfavorable among feature sets, we keep these features in other trials because of their higher precision.

The feature set K (head feature) has the lowest measure in the upper panel, when we compared the performances of feature sets A, C, J, and K. In contrast, when we compared the performances of sets P through V, excluding the head feature (set S) increased the recall and decreased the precision noticeably. Hence, we inferred that the head feature may not be a good indicator for phrase boundary detection but would make the trained model relatively conservative in prediction behavior.

In the upper panel, feature sets A, B, C, F, J, and K included the original token and an extra type of feature, and offered evidence for the contributions of these extra features to identify the agents. In the lower panel, feature sets P, Q, R, S, T, U, and V are similar to each other, and they differed only in one feature. They achieved similar precision rates but different recall rates.

### 5.3 Additional Results of Experiments under Specific Settings

A specific class of sentences may help us achieve better results in the experiments. We chose those sentences that have "agent", one of "expressive-subjectivity" and "objective speech event", and "direct-subjective", and call this condition "explicit expression". In MPQA, we found 4823 instances that satisfy the explicit expression condition, and we used 1447 of them in the test phase. Table 5 shows the experimental results, when we used all sentences and explicit-expression sentences for experiments reported, respectively, in the left and right sides of the table. The statistics clearly indicate that it was easier to find the right semantic tags for the sentences that have explicit expressions.

In Table 6, we compared the performances of a first-order CRF model and a second-order CRF model under explicit expression sentences, but did not find the second-order performance to outperform. A probable reason is that the second-order model is overfitted due to insufficient training data. However, we have not verified this conjecture yet.

We observed that our models performed poorly for the tagging of "target", e.g., in Table 5. Possible reasons of such a phenomenon include (1) that "target" tags rarely appear in the MPQA corpus and (2) that the length of "target" text is longer than other classes. Hence, we removed those "target" tags from our data for better system performance. We compared the results of experiments in which the "target" tags resided and were removed from the data, and the results are reported, respectively in the left and right side of Table 7. In the left side of Table 7, the results for the "target" class are worst. After removing the "target" tags from the data, we obtained better results in all categories, except for the precision of "Objective speech-event".

## 6. EXTRACTING OPINION PATTERNS

We employed the CRF model that we obtained with feature set W in Table 4 to tag the text material that we extracted from financial statements (cf. Section 2). The goals were to extract opinion holders and subjective expressions. We chose set W because of the high precision that it achieved in the experiments with the MPQA corpus. During the preprocessing steps, we replaced personal names, organization names, and locations with PERSON, ORG, and LOC, respectively. The reason for doing so is because we did not need to distinguish entities of these three types in details. We just need to know it was a certain person who did some actions, for instance.

In Panel A of Table 8, we show that the most frequent 8 MWEs for "agent", "direct subjective", and "subjective expression". Since a financial statement is a report of a specific corporation, words "we" and "the company" and "management" referred to the same entity, and they were the most frequent subjects in the sentences. Notice that the most frequent MWEs contained incorrect MWEs, and they are underlined in Table 8.

Providing a procedure to automatically extract subjective multiword expressions in financial statements is a main contribution of this work. We contribute to the literature by extending previous works [19] that only examined unigrams of financial terms. Not only that the most frequent 8 MWEs in "subjective expression" are meaningful, some less frequent MWEs are even more interesting. In Panel B, we show some examples. The expression "scientifically feasible commercially viable opportunity" suggests a positive investment outlook, and the phrase "substantially doubt about its ability to continue as a going concern" indicates risks of bankruptcy in the near future.

Given these MWEs of subjective judgments, we could analyze the financial statements with more advanced techniques. For instance, we can search for sentences in which the agents like "we" or "company" collocated with specific financial MWEs. Statements that meet such criteria shed light on the financial status of the reporting corporations. Table 9 shows some statistics about such statements.

Unlike MPQA, where correct tags were provided, financial statements with appropriate annotations are not available. Hence, we could not conduct large scale experiments to evaluate the performance of using our CRF models to extract opinion patterns from the financial statements.

Nevertheless, we ran an experiment with 100 arbitrarily chosen sentences which we manually annotated the tags. In this experiment, we achieved higher accuracy in labeling "agent" than labeling "direct-subjective" and "expressive-subjectivity". On average, we had 58% of the "agents" labeled correctly. When "direct-subjective" MWEs were actually unigrams, we achieved 83% in accuracy, but when they were 3-grams or longer n-grams, the accuracy dropped significantly. When "expressive-subjectivity" MWEs composed of 5 or fewer words, we achieved 67% in accuracy.

## 7. INTERPRETING THE CORRELATIONS

In this section we present results of our investigation about whether a multiword subjective expression, $w$, was positively or negatively correlated with the *SUEs*, and examine the strength of correlation based on the rank of the correlation, $R(w)$, that we defined in Section 4.4. We also comment on the identified correlations from a financial standpoint. Since our programs could not identify the subjective MWEs perfectly, we used the results of the experiment that employed feature set W, in Table 4, to identify the subjective expressions, and manually selected MWEs to do further experiments.

### 7.1 Empirical Results of Small Dataset

Results in this section are based on the "small" data set that was explained in Section 2. We set the range for separating the increasing and decreasing earnings to 0.5 (i.e., in Equation (3) in Section 4.1), and selected 113 MWEs in this experiment. We summarized some of our observations in Table 10 and divide the table into two major portions. The upper portion shows the observations for the MWEs under investigation, and the lower portions shows statistics about the control variables. The upper portion is divided into five major columns, providing the IDs for the MWEs, the MWEs (denoted by $w$), the coefficients and $p$-values which were estimated by Equation (5) when
and when         , and the rank of influence of $w$ (i.e., $R(w)$). In the columns with headings         and    , we show two numbers. The first number is the coefficient of the MWE in Equation (5), and the second number is the $p$-value for the coefficient. The numbers were calculated with Stata. A correlation was considered significant if its $p$-value was less than 0.05. Coefficients of significant correlations are highlighted in the upper portion.

In Table 10, we can see that relatively soft MWEs were used when earnings were decreasing (i.e.,         ). The word "could" in v20 and v31 and the word "may" in v64 are not definitive. The companies used them to mitigate the implications of the negative words in the financial statements for years with declining earnings.

Recall that the base case in Equation (5) is for        . Hence, we find that it was less likely to observe v100 and v108 in years of increasing earnings (i.e.,         ). In such years, the companies tend not to promise for future success. In contrast, in years with stable earnings, making promises for the future is more often.

Also, the MWE "seriously harmed" carry a very strong and negative message, and is not an ordinary statement to be used by the reporting companies and their auditors in financial statements. Hence, its correlations to either case were insignificant.

Furthermore, we found that the negative MWEs (v20 and v64) are expressed less strongly than the positive MWEs (v100 and v108) given the context of U.S. financial statements. Namely, the words "will" and "always" are stronger than "might" and "could" in subjective strength. The above discussion suggests that the companies would accentuate the positive financial status of the corporations and attempt to use "prudent wording" to obfuscate the negative financial status.

### 7.2 Robustness Checks: More Experiments

We provide robustness checks in this section by expanding our experiments on four aspects. We used 22780 instances in this experiment (cf. Section 2), considered 174 MWEf-idf variables (rather than 113 variables in the previous subsection), and set to 0.5 and 1.0. In an attempt to make the estimated coefficients more reliable, we employed the Huber-White sandwich estimators by activating the "robust" option in Stata. Using the Huber-White sandwich estimators allowed us to consider the issue of the heteroscedasticity and the normality assumption [12, 33].

Table 11 shows the results when we set to 0.5 and experimented with the much larger dataset. Table 11 has the same columns as those of Table 10. The MWEs that had significant correlations in Table 10 and Table 11 are quite different. This is due to the differences in size and population of the data. We used only 1421 instances in the experiment in the previous subsection. Similar to our observation in Table 10, relatively definitive and positive MWEs (v13 and v120) were used with increasing earnings (    ). In contrast, very negative terms (v33 and v77) were seldom used with decreasing earnings (    ). Similar to what we observed in Table 10, "could" (v23) was used when earnings were decreasing.

Table 12 shows the results when we set to 1.0 and experimented with a much larger dataset. Recall that a larger    makes it relatively harder for increasing earnings to qualify as    , i.e., that the companies must have made more money. This could be the reason why the coefficients for v49 and v120 become larger. In addition, we observe that the coefficient for v51 with    , a very negative comment, is very negative as well. The coefficients for v23, v33, and v77 are similar to those in Table 11, suggesting that our findings were quite stable in different settings of the experiments.

It is interesting to observe that the coefficients of v157 (with    ) in Table 11 and Table 12. The cases that belong to    in Table 12 lost more money than those in Table 11, when we increased    from 0.5 to 1.0, cf. Equation (3). The statistics for v157 in these tables show that, when the company was losing an unusual amount of money, managers became more likely to use "will be successful". Promising a good future helps people ameliorate the bad feeling about current (losing) status.

An equally interesting observation is about the phrase - "may be uncollectable" (v90). This MWE is negative in that it suggests the company might manipulate the earnings by selling their products with unusual account receivables that will make it "uncollectable" in the future [29]. The MWE was used with cases when the companies that appeared to make unusual amount of money. When we increased    , instances belonging to the category    represented situations where more money was earned than normal years. Making more money prompted managers to make relatively conservative statement by using "may be uncollectable" more frequently.

The change in the significance of the coefficient for v51 ("doubt about ability to continue as a going concern" indicates that the auditor find some risks of bankruptcy) from Table 11 to Table 12 is also worth noting. The earnings increased much more for instances in the class of     when     is 1.0 than when     is 0.5. Hence, the auditors are less likely to use v51 for instances in      when     is 1.0. It reveals that the auditor would not proclaim negative opinions when the increasing earnings are extremely high.

The drop of $R(w)$ of v13 ("can successfully") from the second to the fifth level, as we increased    , indicates that v13 was not used in extreme cases (     or      when     was 1.0) frequently. Some terms like v88 ("may be successful") seemed to be less closely related to increasing or decreasing earnings. The word "may" not only reduce the strength of subjectivity but also cause uncertainty. This would offset the positive aspect of the word "successful", so the MWE "may be successful" is insignificant.

### 7.3 Strength of the Agent's Attitude in the Financial Statements

Using the method to collect data in Table 9, we found the frequency of a particular combination of an agent and an opinioned MWE. Categorizing words in the MWEs based on their positive and negative implications [20], we analyzed the strength of opinions of different agents when they commented on years with decreasing and increasing annual earnings. Some examples are provided in Table 13.

When *SUE* is positive (     ), the companies were more likely to use strong positive MWEs than to use weak positive MWEs. This is illustrated by the higher correlation of "reasonably assured" with cases in the category of      . Similarly, the auditors used positive words in financial statements for cases in the category of      .

When *SUE* is negative (     ), the companies tend to use words of "weak negative" more often than using "strong negative". The MWE "could adversely affect" appeared more often than the MWE "materially adversely affected" with cases in the category of      . In fact, the correction between "seriously harmed" and the cases in the category of      was insignificant. Similarly, the auditors did not use negative words consistently for cases in the category of      .

## 8. CONCLUDING REMARKS

We present applications of linear-chain CRF models which embrace linguistic features in morphology, orthography, predicate-argument structure, syntax, and simple semantics to identify opinion holder and extract opinion patterns. The best performing CRF models for opinion holder identification achieved 72% in precision when tagging the "agent" class and 81% in token-based accuracy, when we ran experiments with the MPQA corpus.

We extracted opinion holders and subjective MWEs from U.S. financial statements with the best performing CRF model in the MPQA-related experiments. Although we could not extract useful financial MWEs completely automatically, current results provide a list of candidates for us to select with domain knowledge. Being able to semi-automatically find useful financial MWEs from realistic financial statements is a noticeable advance, comparing to papers only used unigrams in the literature [20].

We further applied multinomial logistic regressions to examine the relationships between the MWEs and company's annual earnings of the reporting companies. Results support a common belief that the reporting companies tend to amplify increasing earnings with positive MWEs and mitigate the implications of decreasing earnings with mild and ambiguous MWEs. Since the textual contents in financial statements would deeply influence the creditors and investors in making decisions about the fate of the companies, the companies have motivations to avoid leaking the negative financial status and accen-tuate the positive financial status.

Although we and other researchers, e.g., [6, 9], have attempted to apply modern technologies for natural language processing to extract the subjective expressions, we fail to extract every important financial expressions. Hence, our strategy was to find methods that offer a good tradeoff between the precision and recall rates. The aim was to allow the computer programs to provide a good list of candidate expressions, making it feasible for us to sift the candidates for useful expressions based on domain dependent knowledge.

Some challenges remain in automatic opinion holder identification. First of all, an individual sentence can contain more than one opinion holders, and different opinion holders may express different expressions. Such nesting structures in complex sentences were difficult to handle correctly. Since the opinion agents "Datanalisis' November poll" and "55.3 percent" in Table 1 are nested, the opinion agents and expression matching is a hard nut to crack. Secondly, anaphor resolution is still hard to solve. To avoid the same words appearing repeatedly in statements, pronouns and abbreviations are frequently used. It is our goal to find the identities referred by these substitute words. Finally, co-reference problem is another barrier. How could our algorithms know that "the leading semiconductor company" may refer to "Intel corp."? The problems described above are beyond the prediction competence of our CRF models that capture mainly the syntactic features at the sentence level and at the word level. More algorithmic tools to apprehend semantics are in demand.

An obvious problem arose when we evaluated our CRF models with the MPQA corpus in Section 5. We chose the combinations of features based on our intuition. Employing better methods for feature selection, such as genetic algorithms, may help achieve better results. The features that we listed in Section 3 are mutually dependent. Therefore, using the principle component analysis may help us find the best combination of features for the extractions of MWEs.

We have demonstrated an application of subjective MWEs in financial statements by studying their relationships with the annual earnings. These subjective MWEs can serve as a basis for studying other issues in the financial markets. One can verify whether the opinion patterns are indicative of the future financial performance of the corporations. It is also interesting to examine whether the sentiment of the opinion patterns agrees with the financial ratios reported in the current and future financial statements.

Fig. 1 shows the flow of steps that we discussed in this paper. This procedure can be applied to investigate the relationship between textual contents and quantitative data in different domains. Take the U.S. presidential election for example. It might be interesting to study the relationships between the candidates' speeches and candidates' supports in polls.

## ACKNOWLEDGMENT

This paper is an extended integration of [4] and [5]. The research has been partially supported by research contracts NSC-99-2221-E-004-007 and NSC-97-2221-E-004-007-MY2 of the National Science Council of Taiwan. We are grateful to the anonymous reviewers of previous versions of our papers for their precious comments and suggestions, and we are also grateful to Ms. Chia-Chi Tsai for her timely and instant help to set up the environment for running experiments.

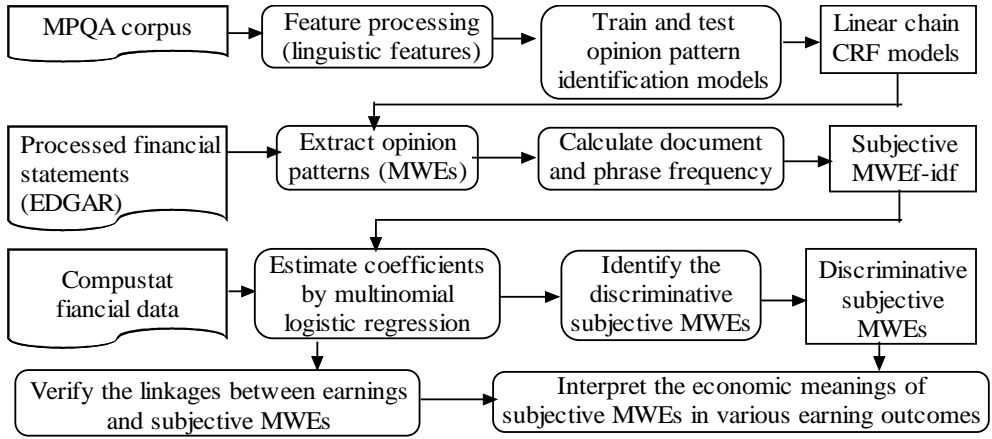

Fig. 1. A block diagram for the work flow of our study

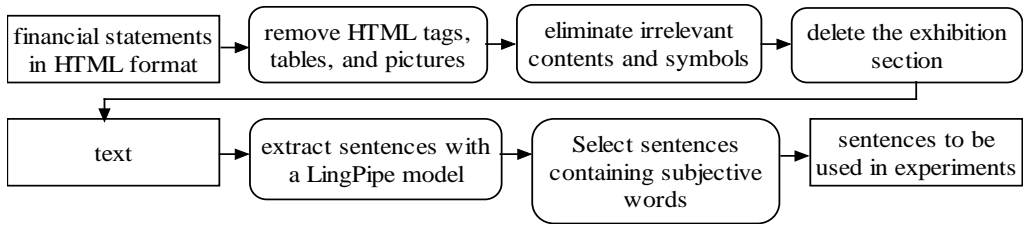

Fig. 2. Extracting sentences from financial statements in the HTML format

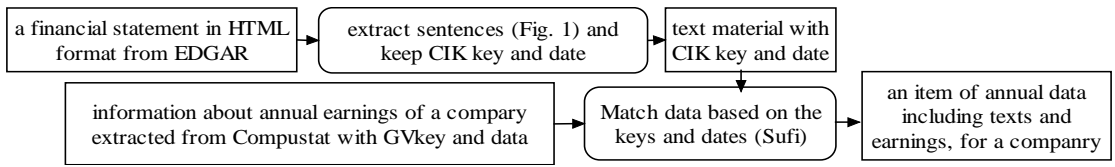

Fig. 3. Matching the sentences in financial statements and the annual earnings for a company

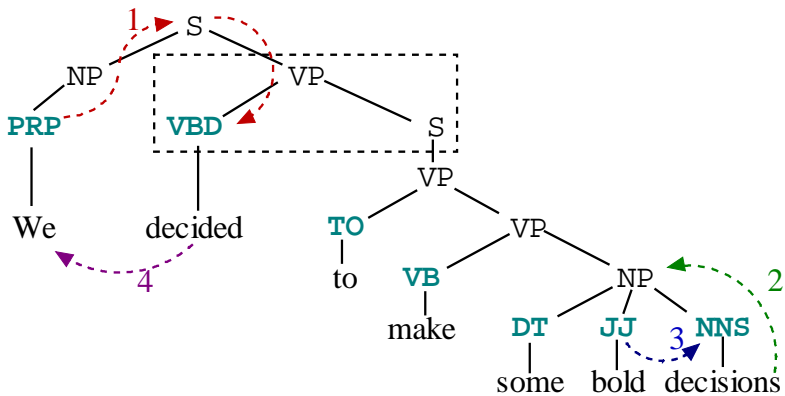

Fig. 4. The syntactic features in a parse tree

Table 1. A sample sentence and its annotations in the MPQA corpus

| According to Datanalisis' November poll, 57.4 percent of those polled feel as bad or worse than in the past and 55.3 percent believe their main problems are in the economic area. | |
|---|---|
| MPQA annotation labels | |
| Opinion holder 1 | Datanalisis' November poll: agent; According to: objective speech event. |
| Opinion holder 2 | 57.4 percent of those polled: agent; feel: direct-subjective; as bad or worse: expressive-subjectivity. |
| Opinion holder 3 | 55.3 percent: agent; believe: direct-subjective; main problems: expressive-subjectivity. |

Table 2. Example sentence expressed in linear CRF data view

| Feature-Level | | | Partially illustrated Features and their values | | | | | | |
|---|---|---|---|---|---|---|---|---|---|
| Token level features | Features of leaf node | f1 | We | decided | to | make | some | bold | decisions |
| | | f2 | we | decide | to | make | some | bold | decision |
| | | f13 | B-NP | B-VP | I-VP | I-VP | B-NP | I-NP | I-NP |
| | | f17 | objective | weak, neutral | objective | objective | objective | strong, positive | objective |
| | | f7 | before | | after | after | after | after | after |
| Phrase level features | Syntactic features of leaf node | f10 | PRP: we | | TO: to | VB: make | DT: some | JJ: bold | NNS: decisions |
| | | f11 | PRP | | TO | VB | DT | JJ | NNS |
| | | f12 | PRP↑NP↑S | | TO↑VP↑ S↑VP | VB↑VP↑VP ↑S↑VP | DT↑NP↑VP↑ VP↑S↑VP | JJ↑NP↑VP↑ VP↑S↑VP | NNS↑NP↑VP↑VP ↑S↑VP |
| | Syntactic features of leaf node's parent | f10 | PRP: we | PREDI- CATE | O | VB: make | O | O | NNS: decisions |
| | | f11 | NP | | O | VP | O | O | NP |
| | | f12 | NP↑S | | O | VP↑VP↑S↑ VP | O | O | NP↑VP↑VP↑S↑ VP |
| Sentence level features | | f8 | active | active | active | active | active | active | active |
| | | f9 | VP→VBD-S | VP→VBD-S | VP→VBD-S | VP→VBD-S | VP→VBD-S | VP→VBD-S | VP→VBD-S |
| IOB (MPQA) | | | B- | B- | B- | I- | | | |
| LABELS (MPQA) | | | agent | direct- subjective | expressive-subjectivity | | | | |

Table 3. Strategy of discriminative MWE identification

| Different significance scenarios | | | R(w) |
|---|---|---|---|
| w **is not** a discriminative indicator of the *SUE* | NSS | NSS | 5th |
| w is a discriminative indicator of the ***negative*** *SUE* (**direct indicator**) | + | - | 1st |
| | + | NSS | 2nd |
| w is a discriminative indicator which is **either** the *positive* *SUE* or **not discriminative** to *SUE* (**indirect indicator**) | - | NSS | 3rd |
| w is a discriminative indicator which is **either** the *negative* *SUE* or **not discriminative** to *SUE* (**indirect indicator**) | NSS | - | 3rd |
| w is a discriminative indicator of the ***positive*** *SUE* (**direct indicator**) | - | + | 1st |
| | NSS | + | 2nd |
| w **is not** a discriminative indicator of the *SUE* | + | + | 4th |
| | - | - | 4th |

Table 4. Results of "agent" identification using different feature sets

| Feature set | | a | p | r | |
|---|---|---|---|---|---|
| Upper Panel | | | | | |
| A | f1+f2(lemma) | 67.64 | 56.60 | 29.48 | 38.77 |
| B | f1+f3~f5(orthographical) | 63.76 | 53.49 | 22.49 | 31.67 |
| C | f1+POS | 64.03 | 66.42 | 16.85 | 26.88 |
| D | f1+POS+f16(NER) | 71.92 | 58.66 | 39.69 | 47.35 |
| E | f1+POS+f15(dependency, dep.) | 71.22 | 62.71 | 42.36 | 50.57 |
| F | f1+f13(base chunk) | 71.16 | 57.05 | 41.83 | 48.27 |
| G | f1+f2+POS+f13+f15~f17 | 66.01 | 66.67 | 25.35 | 36.74 |
| H | f1~f5+POS+f13+f15~f17 | 65.77 | 69.77 | 19.22 | 30.14 |
| I | f1+f2+POS+f13+f14~f17 | 65.79 | 69.09 | 27.92 | 39.77 |
| J | f1+f11(phrase type) | 70.89 | 69.36 | 17.32 | 27.72 |
| K | f1+f10(head) | 70.67 | 27.07 | 4.27 | 7.37 |
| L | f1+f11+f12(phrase type and head) | 70.64 | 65.09 | 16.99 | 26.94 |
| Lower Panel | | | | | |
| M | f1+f12(only path) | 71.14 | 62.52 | 15.07 | 24.29 |
| N | f1+f10+f12(path and head) | 71.04 | 60.05 | 16.59 | 26.00 |
| O | f1+f10+f11(phrase type and path) | 71.02 | 68.92 | 21.14 | 32.36 |
| P | f1,f2,f6~f19(dep. excluded) | 70.71 | 64.01 | 50.60 | 56.52 |
| Q | f1,f2,f6~f19(NER excluded) | 70.88 | 69.91 | 35.91 | 47.45 |
| R | f1,f2,f6~f19(path excluded) | 71.23 | 68.62 | 35.73 | 46.99 |
| S | f1,f2,f6~f19(head excluded) | 71.04 | 64.69 | 48.25 | 55.27 |
| T | f1,f2,f10~f17(predicate excluded) | 71.04 | 69.40 | 38.02 | 49.12 |
| U | f1,f2,f6~f19(path, dep. excluded) | 70.64 | 67.90 | 32.15 | 43.63 |
| V | f1,f2,f6~f19(full) | 70.93 | 69.96 | 36.45 | 47.93 |
| W | f1,f2,f6~f19(no "target" label) | 76.97 | 70.84 | 38.28 | 49.70 |

Table 5. Performance with explicit expression sentences

| Annotation labels | All sentences | | | Explicit expression | | |
|---|---|---|---|---|---|---|
| Feature set | G: (f1+f2+POS+f13+f15~f17 with "target" label) | | | | | |
| % | p | r | | p | r | |
| Agent | 66.67 | 25.35 | 36.74 | 64.34 | 43.61 | 51.98 |
| Obj.speech-event | 36.89 | 9.87 | 15.57 | 44.12 | 36.36 | 39.87 |
| Direct-subjective | 44.61 | 14.75 | 22.17 | 51.09 | 32.99 | 40.09 |
| Expressive-sub. | 7.65 | 0.80 | 1.46 | 27.67 | 10.49 | 15.21 |
| Target | 0.65 | 0.10 | 0.17 | 10.19 | 5.37 | 7.03 |
| Other | 37.43 | 51.23 | 43.25 | 48.29 | 57.42 | 52.46 |
| Average | 37.72 | 44.48 | 40.82 | 48.09 | 52.20 | 50.06 |
| Token accuracy | 66.01 | | | 68.79 | | |

Table 6. Performances achieved by the 1st and 2nd order CRF

| Annotation labels | First-order CRF | | | Second-order CRF | | |
|---|---|---|---|---|---|---|
| Feature set | W (explicit expression and without "target") | | | | | |
| % | p | r | | p | r | |
| Agent | 70.54 | 41.95 | 52.61 | 69.29 | 39.17 | 50.05 |
| Obj. speech-event | 40.00 | 5.88 | 10.26 | 46.15 | 5.43 | 9.72 |
| Direct-subjective | 44.22 | 18.38 | 25.97 | 45.07 | 17.20 | 24.90 |
| Expressive-sub. | 18.06 | 3.43 | 5.76 | 8.15 | 1.45 | 2.46 |
| Other | 32.97 | 42.12 | 36.99 | 30.68 | 39.62 | 34.58 |
| Average | 34.12 | 39.09 | 36.43 | 31.78 | 36.68 | 34.05 |
| Token accuracy | 73.52 | | | 72.31 | | |

Table 7. Performance with and without "target" label

| Annotation labels | With target label | | | Without target label | | |
|---|---|---|---|---|---|---|
| % | p | r | | p | r | |
| Agent | 63.68 | 27.99 | 38.89 | 72.29 | 30.81 | 43.21 |
| Obj.speech-event | 42.86 | 1.07 | 2.09 | 29.07 | 4.50 | 7.80 |
| Direct-subjective | 46.69 | 9.11 | 15.25 | 56.23 | 11.96 | 19.72 |
| Expressive-sub. | 14.18 | 1.74 | 3.10 | 27.67 | 5.01 | 8.48 |
| Target | 4.88 | 0.19 | 0.36 | - | - | - |
| Other | 40.86 | 52.45 | 45.94 | 46.98 | 56.37 | 51.25 |
| Average | 41.03 | 47.81 | 44.16 | 47.19 | 52.4 | 49.66 |
| Token accuracy | 75.08 | | | 81.27 | | |

Table 8. Opinion patterns extracted from 10-K filings

| Panel A   Top 8 frequent phrases | | | | | |
|---|---|---|---|---|---|
| Agent | Freq. | Direct subjective | Freq. | Subjective expression | Freq. |
| We | 8249 | believe | 5352 | may not be able | 140 |
| the company | 1840 | agree | 1274 | may not be recoverable | 137 |
| the ORG. | 948 | expect | 888 | reasonably assure | 62 |
| management | 606 | cannot assure you | 635 | may be impaired | 55 |
| the | 408 | intend | 546 | substantial doubt | 51 |
| It | 394 | do not believe | 538 | would not be able | 49 |
| company | 328 | provide | 466 | may not be successful | 48 |
| PERSON | 249 | determine | 346 | would become exercisable | 45 |
| Panel B   Some of the other frequent phrases | | | | | |
| the plaintiff | 76 | anticipate | 169 | scientifically feasible commercially viable opportunity | 22 |
| the executive | 71 | deny | 143 | could adversely affected | 19 |
| the debtor | 35 | conclude | 130 | could significantly reduce our revenue | 16 |
| the credit agreement | 31 | violate | 19 | substantially doubt about its ability to continue as a going concern | 15 |

Table 9. Opinion patterns combination in sentences

| Opinion patterns combination | Total freq. | Document freq. |
|---|---|---|
| company AND "will be able" | 110758 | 693 |
| company AND "successfully" | 95549 | 738 |
| company AND "adversely affected" | 115858 | 518 |
| we AND "may not be able to" | 73531 | 428 |
| we AND "may be impaired" | 5332 | 29 |
| we AND "seriously harmed" | 2517 | 22 |

Table 10. Empirical results of small dataset

| ID | Explanatory variables (w) | | | | R(w) |
|---|---|---|---|---|---|
| | Significant coefficients of the MWEf-idf variables | | | | |
| v1 | adequately | | 2.086 ; 0.267 | **4.586 ; 0.004** | 2nd |
| v7 | can be no assurance that company will be able | | **4.482 ; 0.034** | 3.317 ; 0.095 | 2nd |
| v20 | could be adversely | | **11.19 ; 0.015** | -1.931 ; 0.708 | 2nd |
| v31 | could seriously | | **3.267 ; 0.011** | 0.245 ; 0.878 | 2nd |
| v64 | may not be able | | **9.755 ; 0.045** | 3.317 ; 0.470 | 2nd |
| v100 | will always be able to | | -3.568 ; 0.325 | **-10.53 ; 0.020** | 3rd |
| v108 | will successfully | | -2.298 ; 0.237 | **-3.792 ; 0.041** | 3rd |
| | Insignificant coefficients of the MWEf-idf variables | | | | |
| v81 | seriously harmed | | -0.984 ; 0.712 | 2.222 ; 0.314 | 5th |
| v59 | may be successful | | 0.398 ; 0.894 | -5.054 ; 0.133 | 5th |
| | Control variables | | | | |
| BM | 0.224 ; 0.008 | -0.128 ; 0.104 | ROE | -0.174 ; 0.000 | 0.046 ; 0.141 |
| Size | -0.051 ; 0.218 | -0.008 ; 0.849 | Accrual | -0.000 ; 0.415 | 0.001 ; 0.076 |
| Dividend | 13.838 ; 0.126 | -1.031 ; 0.924 | Asset growth | 0.001 ; 0.714 | -0.048 ; 0.236 |
| Bankruptcy Z-score | -0.0024 ; 0.501 | -0.002 ; 0.442 | Lag SUE | 0.000 ; 0.135 | -0.001 ; 0.037 |

Table 11. Expanded experiment with =0.5

| ID | Explanatory variables (w) | | | R(w) |
|---|---|---|---|---|
| | Significant Correlations | | | |
| v13 | can successfully | 1.003 ; 0.406 | **2.557 ; 0.021** | 2nd |
| v120 | reasonably assured | -0.346 ; 0.476 | **0.780 ; 0.044** | 2nd |
| v49 | did not contain an adverse opinion or disclaimer of opinion | -3.729 ; 0.416 | **6.766 ; 0.029** | 2nd |
| v23 | could adversely affect our business | **1.135 ; 0.043** | 0.582 ; 0.261 | 2nd |
| v33 | could be materially adversely affected | **-1.111 ; 0.019** | -0.228 ; 0.621 | 3rd |
| v77 | may adversely impact our business | **-8.827 ; 0.005** | -1.003 ; 0.602 | 3rd |
| v157 | will be successful | **0.797 ; 0.042** | 0.333 ; 0.382 | 2nd |
| v90 | may be uncollectible | 7.783 ; 0.135 | **12.42 ; 0.027** | 2nd |
| | Insignificant Correlations | | | |
| v88 | may be successful | -4.291 ; 0.238 | -5.740 ; 0.125 | 5th |
| v122 | seriously harmed | -0.191 ; 0.711 | 0.845 ; 0.070 | 5th |
| v51 | doubt about ability to continue as a going concern | -3.964 ; 0.499 | 0.339 ; 0.954 | 5th |

Table 12. Expanded experiment with =1

| ID | Explanatory variables (w) | | | R(w) |
|---|---|---|---|---|
| | Significant Correlations | | | |
| v120 | reasonably assured | -0.341 ; 0.512 | **1.181 ; 0.007** | 2nd |
| v49 | did not contain an adverse opinion or disclaimer of opinion | 1.758 ; 0.640 | **7.389 ; 0.017** | 2nd |
| v23 | could adversely affect our business | **1.287 ; 0.038** | 0.001 ; 0.999 | 2nd |
| v33 | could be materially adversely affected | **-1.649 ; 0.004** | 0.067 ; 0.898 | 3rd |
| v77 | may adversely impact our business | **-7.300 ; 0.037** | -2.962 ; 0.276 | 3rd |
| v51 | doubt about ability to continue as a going concern | -0.381 ; 0.942 | **-94.45 ; 0.000** | 3rd |
| v157 | will be successful | **1.139 ; 0.008** | 0.180 ; 0.687 | 2nd |
| v90 | may be uncollectible | 6.307 ; 0.163 | **15.49 ; 0.002** | 2nd |
| | Insignificant Correlations | | | |
| v13 | can successfully | 1.170 ; 0.360 | 2.116 ; 0.096 | 5th |
| v88 | may be successful | -4.867 ; 0.316 | -4.318 ; 0.334 | 5th |
| v122 | seriously harmed | 0.823 ; 0.153 | 1.086 ; 0.059 | 5th |

Table 13. Economic meaning of the subjective MWEs

| Opinion holder | Subjective MWE ($w$) | Opinion polarity | $R(w)$ | Condition |
|---|---|---|---|---|
| The Company | reasonably assured | strong positive | $2^{nd}$ | |
| | may be successful | weak positive | $5^{th}$ | |
| | could adversely affect our business | weak negative | $2^{nd}$ | |
| | could be materially adversely affected | | $3^{rd}$ | |
| | seriously harmed | strong negative | $5^{th}$ | |
| Auditor | did not contain an adverse opinion or disclaimer of opinion. | positive | $2^{nd}$ | |
| | doubt about ability to continue as a going concern. | negative | $3^{rd}$ or $5^{th}$ | |